\documentclass[11pt,a4paper]{article}
\usepackage[implicit]{hyperref}
\usepackage{times}
\usepackage{latexsym}
\usepackage{booktabs}
\usepackage{url}
\usepackage{listings}
\usepackage{color}
\usepackage{graphicx}
\usepackage{amsmath}
\usepackage{cleveref}
\definecolor{dkgreen}{rgb}{0,0.6,0}
\definecolor{gray}{rgb}{0.5,0.5,0.5}
\definecolor{mauve}{rgb}{0.58,0,0.82}

\title{Semantic Motion Correction Via Iterative Nonlinear Optimization and Animation}

\author{
  Sairamvinay Vijayaraghavan,
  Jinxiao Song,
  Wan-Jhen Lin,
  Michael J Livanos
  }

\date{}
\begin{document}
\maketitle
\section{Introduction}
In soccer games, the goalkeeper tries their best to guard the ball against going into the goal post in order to prevent the opponent from scoring, however they often make incorrect judgment calls on the prediction of the ball's direction, resulting in a goal against their team. There are many factors involved when we consider what can affect soccer’s movement. For example, the ball may move in one direction, but the goalkeeper moves in the opposite direction. Even the goalkeeper may detect the correct direction but miss the ball by some offset in the position of the ball (due to insufficient diving). We propose an end-to-end method of animating the goalkeeper and the ball's motion and correcting it such that it retains much of its semantic qualities, but the goalkeeper is able to accomplish their task. This can be interesting for coaches and players to view, as producing an animated video with similar but corrected motion has potential pedagogical implications. Further, fans might be also interested in seeing 'what if' scenarios for real games. While in this paper we exclusively focus on the task of soccer penalty kicks, we believe this method can be applicable to similar physical tasks.

We have two important parameters to capture for resolving this problem: the ball and the goalkeeper motion. In order to capture the goalkeeper’s motion, we use pose detection techniques to find a set of key-points at different joints which forms a skeleton with a human in the loop system ensuring that the correct player is being isolated, and detection is accurate. Similarly, for ball detection, we use object detection coupled with manual detection of the ball to track the soccer ball.

With both the goalkeeper's position (represented as the skeleton) and the ball, we can create an animation for their motion by using the key-frames, and apply our optimization technique on each of the frames.

\begin{figure}
    \centering
    \includegraphics[scale=0.11]{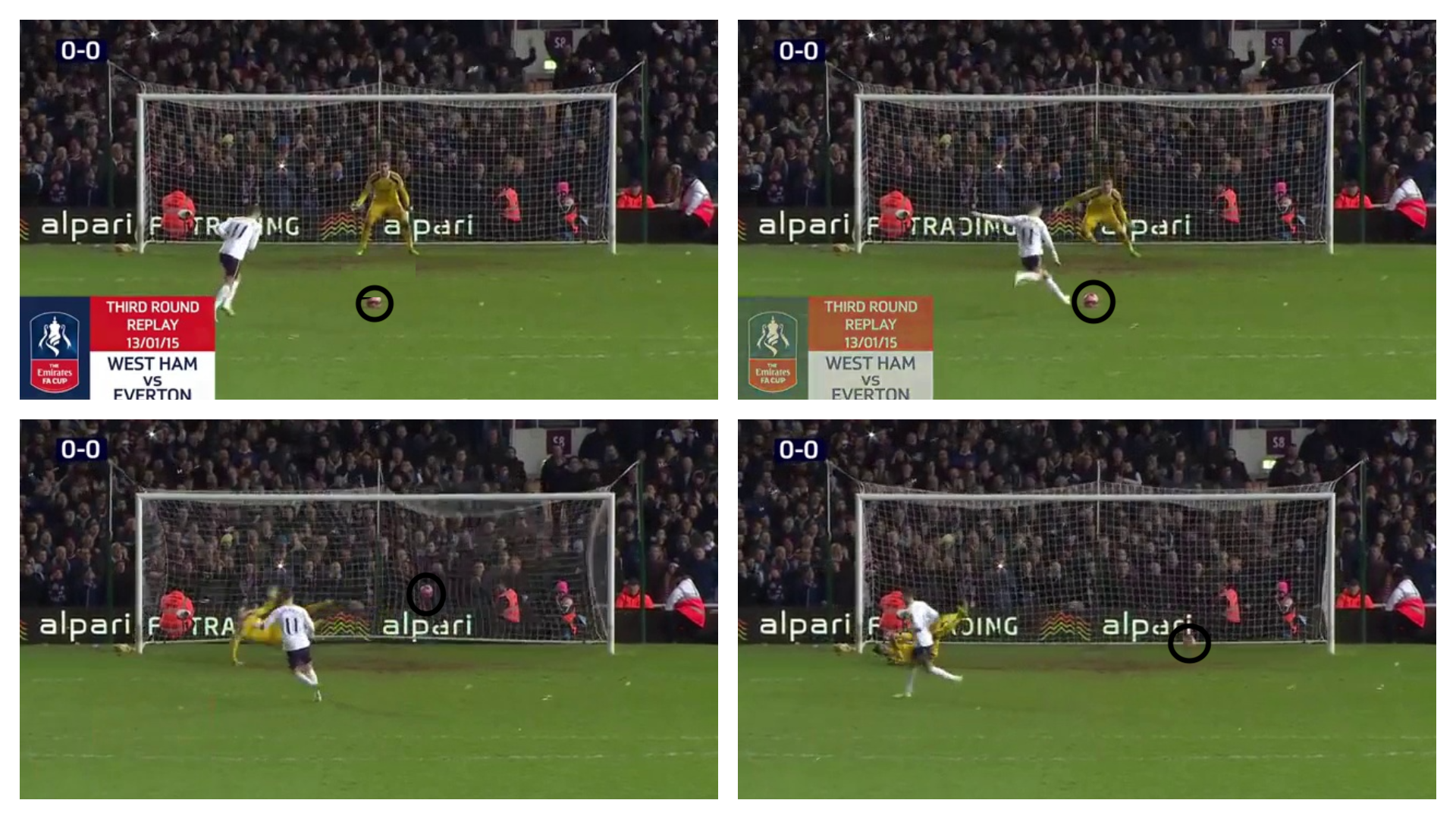}
    \caption{Example of a 'hit'. The goalkeeper fails to stop the goal.}
    \label{fig:hit_soccer_image}
\end{figure}

We handpick a collection of penalty kick videos in which the goalkeepers fail to stop the goal. Our novel dataset, with both the original videos and processed keyframes, along with our code, can be accessed on our public online repository\footnote{https://github.com/Sairamvinay/Football-detection}.

\begin{figure}
    \centering
    \includegraphics[scale=0.11]{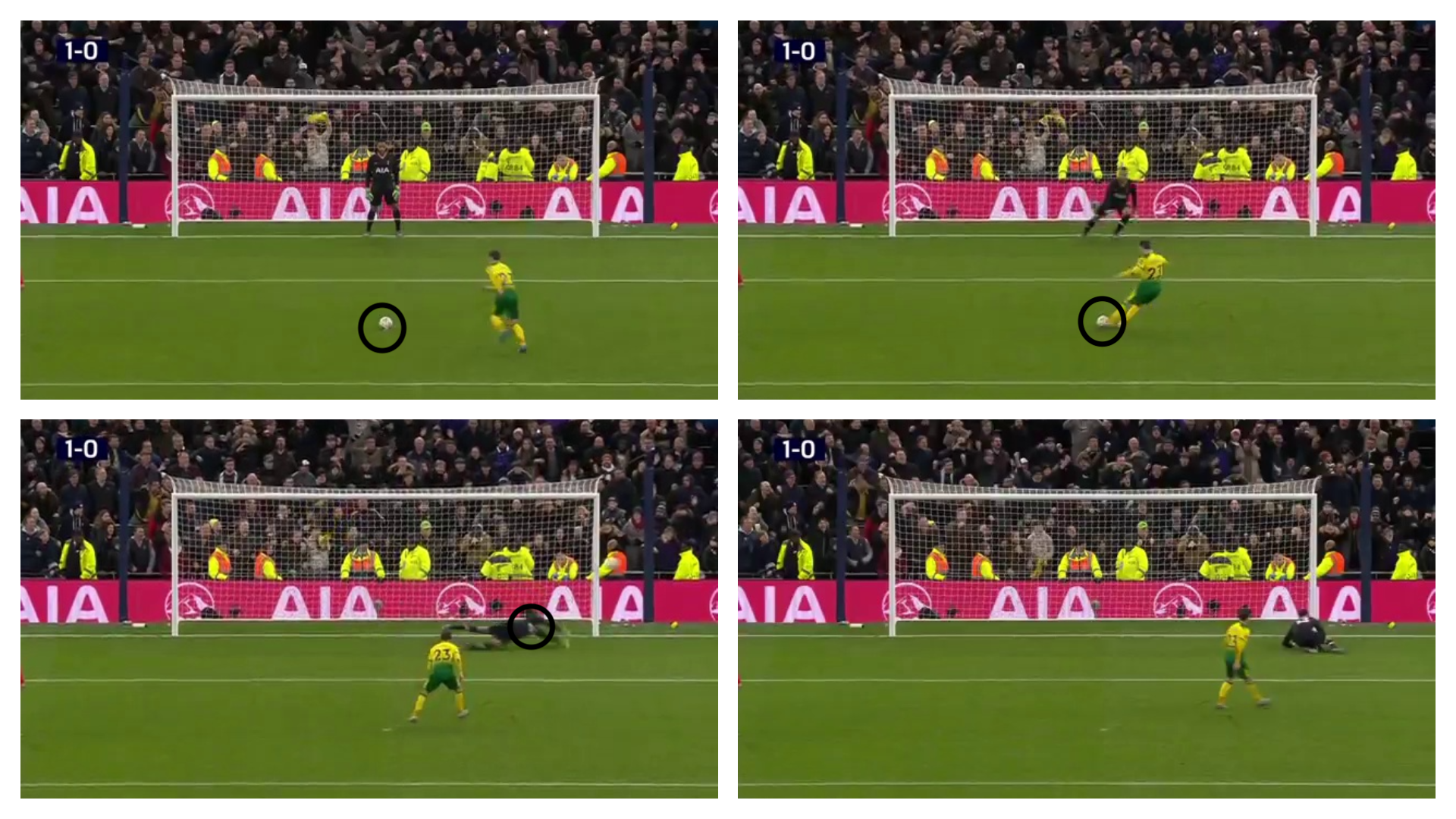}
    \caption{Example of a miss. The goalkeeper is able to block the shot.}
    \label{fig:miss_soccer_image}
\end{figure}

\section{Related Work}

There are several existing methods to tracking human motion and output the positions of their joints. VNect \cite{VNect_SIGGRAPH2017} successfully tracks people with a single RGB camera and can output animated motion. Other work, such as \cite{andriluka2008people} use a pedestrian detector to capture the motion of multiple people. Recently, more and more techniques have been used with machine learning and deep learning. GoodPoint\cite{belikov2020goodpoint} can detect points and generate keypoint descriptors on images. However, all of them list blurry frames and fast motion as a limitation. Further, because there may be many individuals in the frame (the kicker, referee, fans, camera crew), existing techniques are not sufficient on their own to ensure we are capturing the motion accurately for the correct person. We therefore implement a human in the loop system, in which a proposed skeleton is generated for a frame, and we must accept or reject the candidate solution. The next candidate solution will be based off of the prior, to help mitigate those issues in the future. 

In order to deal with these challenges, we elect to base our keypoint detection algorithm from OpenPose\cite{openpose}. Not only is this one of the most robust method of single camera 2D keypoint detection, but it is also sufficient to handle an arbitrary number of people. Specifically, we leveraging their proposal network, and changing it so that it produces candidate solutions for the human to confirm.

For ball detection, there is research that detailed the method to capture the motion of the ball. However, most of these solutions cannot detect very fast motion correctly. Although \cite{lin2011high} develops a high-speed 3D motion capture system, it requires higher resolution. In soccer games, the ball moves very quickly, and the video quality is not as high as other games.

One technology that one might be drawn to to solve this problem is the idea of counterfactual instances\cite{IML}. This explainable artificial intelligence (XAI) technique leverages the idea of counterfactual reasoning in humans in which someone postulates how an event would have ended differently if the preceding conditions were different. For example, one might reason that they would have passed a test if they had studied more, with 'studying more' being the counterfactual event, and passing the test the desired outcome. In XAI, this is formalized by asking what the minimum change needed to create a virtual instance $X'$ to an instance $X$ such that a machine learning model $f$ changes its output from $y$ to $y'$. That is, they find $argmmin_d f(X+d)=y'$.\cite{cf1}

While this kind of work may sound appealing for the task, we cannot apply this work for our purposes because the notion of finding the argmin of a perturbation does not translate well for time series data such as a series of keypoints. One could argue that this problem reduces to a representation learning problem, however we also have to consider the fact that the only part of the instance we can change is the motion of the actor, and not the ball. This problem is further complicated by the fact that counterfactuals tend to not produce interpretable results for images, let alone video\cite{cf2}. While there may be a technique to leverage counterfactuals for this problem, this remains a non-trivial open research question.

\section{Dataset}

We had to collect the soccer match penalty shootout videos from various tournaments such as FIFA, FA Cup, Asia Cup, Spanish Super Cup, and NCAA, etc. We define one sample as a single front view video that shows the soccer ball's motion towards the goal post and the goalkeeper’s motion. This view allows us to simplify the problem into a 2D animation. We then label our sample set into two classes: hits and misses. Hit defines the case when the goalkeeper fails to stop the ball (the wrong action taken in terms of position and maybe direction), and a miss means that the goalkeeper stops the ball from reaching the goal post (the correct action taken in terms of the position as well as the direction). Overall, we have 256 hit and 74 miss samples from which we handpicked certain high quality and well-defined videos for our analysis. We only collect front view videos to simplify our detection for the ball and the goalkeeper. We collect so much data because, originally we planned on training a recurrent neural network to create the final position of the goalkeeper and performing optimization around that virtual goal, however we found that simple techniques which do not require machine learning were perfectly sufficient for this task. The use of machine learning could be interesting for future work, however, so we include all of our data (both the raw videos and skeletons) in the repository.

There are several pre-processing steps that we performed after we collected the videos. Since each video has different amounts of frames, we wanted to avoid the bias of the video length’s impact on the motion. Therefore, we decided to split each video into 10 select frames which are evenly separated in time. Then, we started to implement our motion detection of the ball and the goalkeeper. The goalkeeper has been tracked with 13 key joint points: the head, left shoulder, right shoulder, left elbow, right elbow, left wrist, right wrist, left hip, right hip, left knee, right knee, left ankle, and right ankle. Details on pose detection are provided in Section \ref{sec:posedetection}.

The soccer ball is tracked within each frame based on the coordinates of the radius of the soccer ball. In the end, we record the frame size that helps to scale the motion coordinates with respect to the frame height and weight in the animation part. Details on this method are provided in Section \ref{sec:balldectect}.

\section{Methodology}
\begin{figure}[!ht]
    \centering
    \includegraphics[scale=0.5]{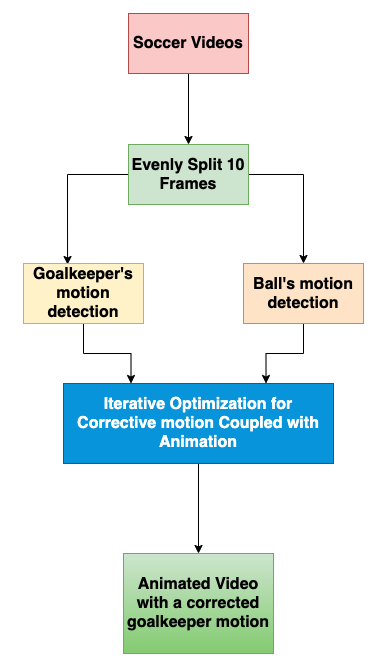}
    \caption{The overall pipeline of our project. We split the video into ten evenly spaced frames, and for each of the frames we find the skeleton of the goalkeeper as a set of keypoints, and mark the ball. After the video is processed, we put feed the data into the iterative optimization procedure, and animate the output.}
    \label{fig:pipeline}
\end{figure}

\subsection{Pose Detection}
\label{sec:posedetection}
Our process of creating the animation starts by analyzing the ten evenly split frames. In each frame, we need to obtain the key points of the goalkeeper in order to transform these key points into human motion animation. For every video, leveraging OpenPose trained on the COCO dataset, we read in a video and evenly (based on the time gap between the consecutive frames) split it by 10 frames.

Before any particular human detection, we had to perform slight pre-processing on the frames. We apply a Gaussian Blur on the frame image which uses a low-pass filter that removes the high-frequency components in the graph. With this step, we can have a smoother image without noise and unnecessary details.

In the next step we use OpenPose to find all the key point joints of the goalkeeper. Among all candidates, our target is to choose the goalkeeper correctly. We use a contour approximation method to detect all points, and then identify possible joints positions of the candidates. Using the Open CV library as the main tool, contours are built by finding all the continuous points along the boundary. From there, we are able to obtain a shape of the object boundary which has the same color or intensity. This technique has been used often for detecting and recognizing objects.

In the next step, we chain all the boundaries by compressing them into one point, which we determine as the key point. Except for the first frame, we will calculate the distance between the point of the previous frame and the point of the current frame. By comparing the result of distance, we are more likely to predict the possible points automatically, however we have a human check if the pose looks correct, and will query the network again if the pose is not correct. There is also manual supervision coupled with this algorithm in case the algorithm fails to detect the key points appropriately.

In addition to getting these coordinates, we decided to normalize our coordinates into a range [-1,1] based on the frame’s height and width, with respect to the axes where the origin was kept as the center of the frame (the $x$ coordinate as the half of the width and the $y$ coordinate as the half of the height). We also implement this normalization technique for the ball detection in order to maintain the same scaling for the animation rendering.
\subsection{Ball Detection}
\label{sec:balldectect}
While performing ball detection, we encountered certain challenges owing due to the combined effect of the low quality of the video and the high speed at which the soccer ball moves causing the supervised ball detection to become infeasible using complicated learning algorithms such as a neural network. This has led us to utilize a much simpler method for detecting the ball’s positional motion.

In order to overcome this challenge and correctly obtain all the ball’s position coordinates in a video, we choose to employ a manually supervised technique. We read in each of the ten evenly split frames and detect the ball coordinates manually. Then, we store the 2D positional coordinates of the ball from the frames which we required for the animation part. Again, we used the same scaling method described above for capturing the goalkeeper’s motion. We chose to capture the ball’s motion separately with respect to the goalkeeper’s motion. This helps us to effectively capture the ball’s trajectory as well as its projection towards the goalpost.

\begin{figure}
    \centering
    \includegraphics[scale=0.07]{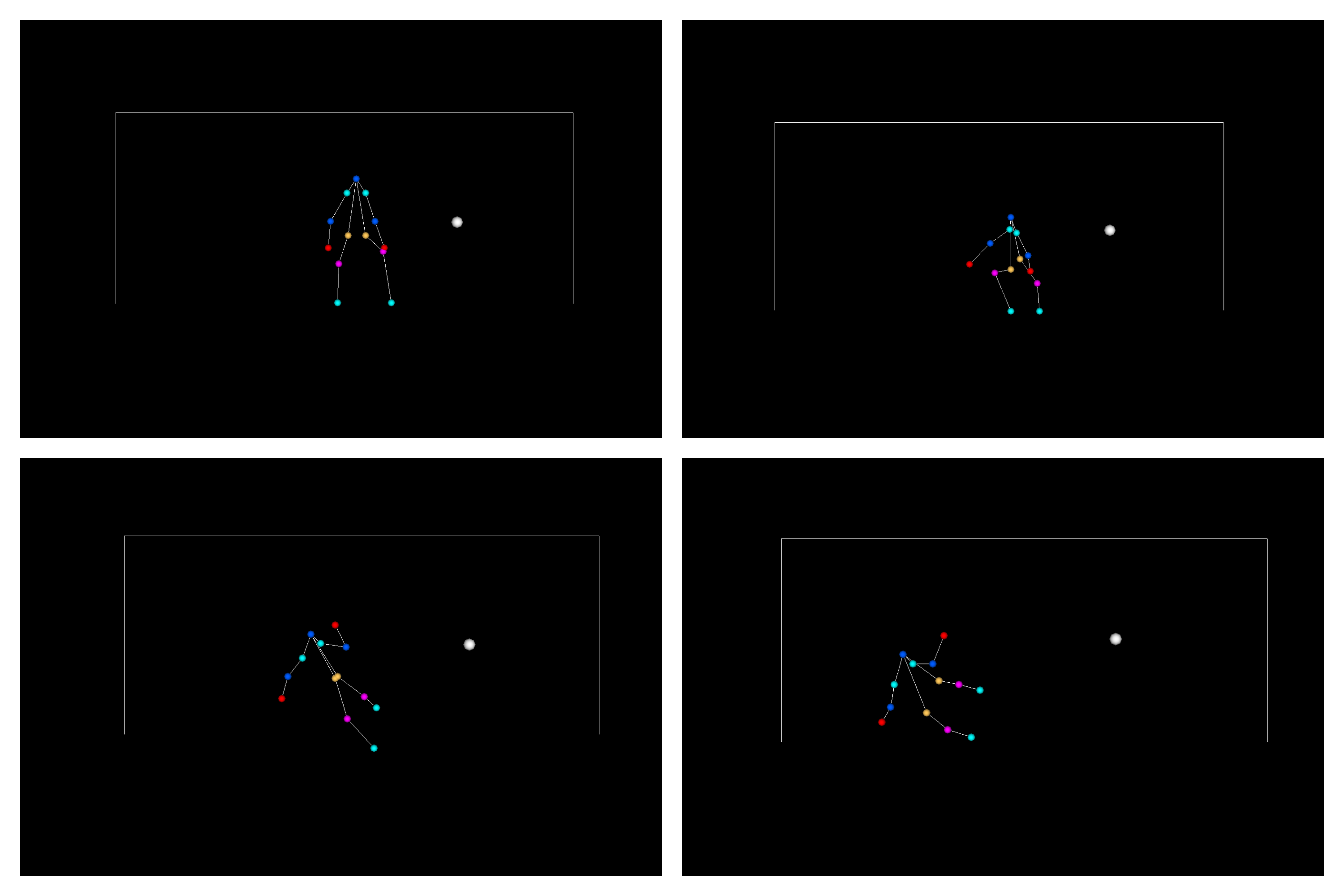}
    \caption{Animation of the sample with the goal frame, the goalkeeper and the ball}
    \label{fig:hit animation}
\end{figure}

\subsection{Optimization}
To create smooth animation, we use an iterative optimization approach to create robust and fluid animation for correcting the motion of the goalkeeper. The approach can be broken down into two steps: end frame generation and optimization via quadratic interpolation. We repeat the second step until motion is fluid. In the first step, a desired final frame position is calculated by selecting a joint to block the ball and superimposing the joint over the ball at the correct time-step. This can be done automatically by finding the closest joint or specified by the user which was used in this project.

We categorize our samples into two types based on the trajectory directions. First, the ball and goalkeeper go in different directions. Second, the goalkeeper goes in the right direction as the ball. In the case where the ball is on the different side of the goal than the goalkeeper, the goalkeeper’s position is flipped around the line created from their head and the hip. Then, the goalkeeper’s blocking joint is superimposed over the ball. This virtual frame is hereby referred to as the virtual goal frame.

Now, for every frame in the video, we generate a new virtual frame by performing quadratic interpolation for every joint in the skeleton, using the corresponding joint from the closest two frames and the goal to perform optimization. We repeat this for every joint in the skeleton, and every frame except for the virtual goal frame and the first frame. Formally, we perform:

$\forall f \in frames, f \neq 0 \& f \neq G, \forall i \in joints:$

\begin{equation}
    \begin{split}
        f_{2}(x_{i}) &= \frac{(x_{i} - x_{G})(x_{i} - x_{i-1})}{(x_{i+1} - x_{G})(x_{i+1} - x_{i-1})} y_{i+1} \\
        &+ \frac{(x_{i} - x_{i+1})(x_{i} - x_{i-1})}{(x_{G} - x_{i+1}) (x_{G} - x_{i-1})} y_{G} \\
        &+ \frac{(x_{i} - x_{i+1}) (x_{i} - x_{G})} {(x_{i-1} - x_{G})(x_{i-1}) - x_{i+1})}y_{i-1}
    \end{split}
\end{equation}   

Where $x_{i-1}$ is the joint at the previous frame, $x_{i+1}$ is the joint at the next frame, $x_{G}$ is the joint at the goal, and $x_{i}$ is the joint at the frame we are optimizing. For the special cases of $i=10$ or $i\pm1 = G$, we use the two nearest neighbors to $i$ rather than $i\pm1$.

\begin{figure*}
    \centering
    \includegraphics[scale=0.5]{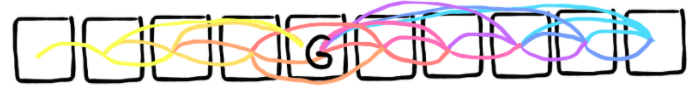}
    \caption{Color diagram detailing the iterative approach. Each frame is computed as the quadratic interpolation between the goal and it's two nearest neighbors. In the figure, each frame has a curve in its own color such that the curve starts at the frame being computed, and ends at one of the frames used for interpolation. For isntance, the penultimate frame (dark blue curves) has curves which extend to the the virtual goal frame and the two frames surrounding it. $G$ represents the virtual goal frame, which in this example is at index 4. Both the virtual goal frame and the start frame remain constant throughout iterations.}
    \label{fig:pipeline2}
\end{figure*}

Keeping the goal in every step of the interpolation helps ensure that there will be a smooth transition to the virtual goal frame, while keeping the start frame constant and using two nearby points which are heavily influenced by the original video helps ensure motion remains semantically similar to the original video. The result is motion that moves smoothly to the virtual goal frame, while maintaining much of the structure of the original video.


\subsection{Animation}
The biggest portion of our project came in designing the animation for the corrected motion. We use the “Visualization Toolkit (VTK)” in Python for simulating all our motions. This library not only allowed us to increase the interactivity amongst users but also provided scope for us to leverage a better user experience.

We had set our render window to a size of 2500 x 1200. We chose this size so that it can accommodate the average frame size of any video from our collection so that we can keep it scaled onto a larger window. For example, certain videos have frames of size as 1280 x 720 while some videos had the frame size as 640 x 360. Videos with a larger frame size than 1280 x 720 were resized to this size so that we maintained uniformity. So, we chose a window size for animation based on the maximum size of the frame and also a convenient size for displaying the characters.

Animating the goal post becomes the biggest challenge. From a trail of errors based on different solutions, we have addressed the challenge by internally updating the goal post with respect to the goalkeeper’s coordinates. From our observation, all the videos had demonstrated that the goalkeeper would always be in a steady position in the first frame and the goalpost would be always behind the goalkeeper. Then, we picked the left and right shoulders of the goalkeeper from the first frame of motion. From there, we set the width of the goalpost as a sum of the difference in the x-coordinates of the shoulders and twice the sum of an offset (set default to 5). Similarly, we set the height of the goalpost as 5 as default. \\\\
 Next, we had to scale the ball’s coordinates as well as the joint’s coordinates. Since we obtained all these values within a range [-1,1], we have to transform these coordinates into the render’s space and hence we applied a unique method of scaling. We decided to obtain a magnification twofold: one to scale the coordinates according to the current render window size and another was a custom magnification we applied for an enlarged view of the goalkeeper as well as the ball. We set the ball’s color to be white for visualization. During our scaling process, we had some coordinates mapped with a negative sign (this is mainly because some coordinates were found mostly in the bottom half of the frame) and this led to our motion being captured in the reverse direction. So, in order to fix this issue, we obtain the absolute value of the coordinates before scaling. Also, for each of the coordinates, we chose to omit the last frame since it did not capture the entire motion and was rather only capturing the ball and goalkeeper’s motion after the ball was blocked/scored. So, we deliberately skipped just the last frame’s coordinates for a cleaner animation.

\section{Result}
\label{sec:results}
We apply our approach on three of the collected videos, each representing a different kind of mistake the goalie can make. For all of these techniques, the optimization took less than one second to complete on an Intel Core i7 2.6 GHz CPU.

\begin{figure*}
    \centering
    \includegraphics[scale=0.7]{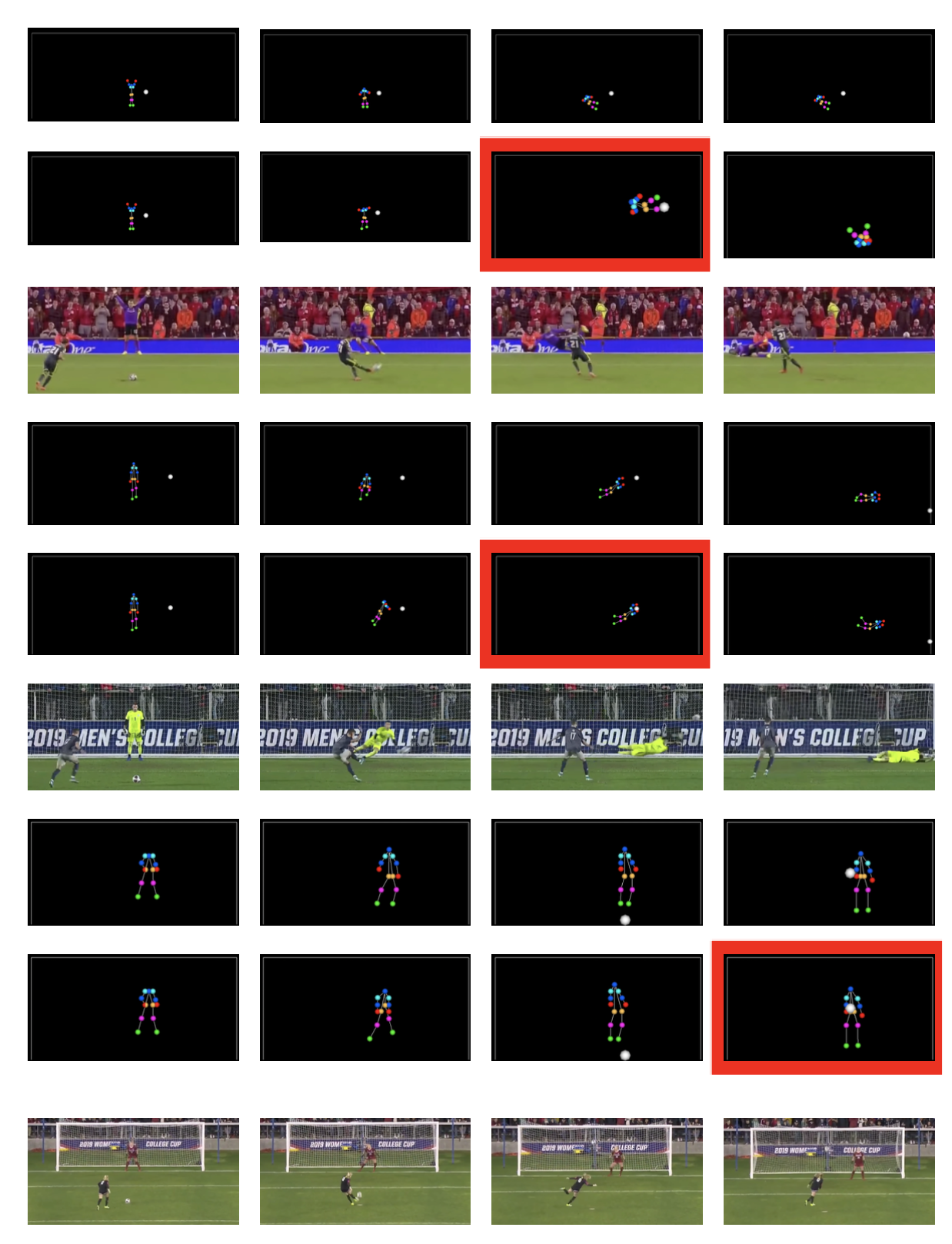}
    \caption{Results from the applied optimization. Every group of three rows represents one video, with the first of those three rows being the detected skeleton, the second the corrected skeleton, and the third being the video it was generated from. Every column is a frame, and four frames are shown (rather than all ten frames produced) to enhance readability. The virtual goal frame is marked in red, and three examples are shown. Video of the skeleton animation is provided in the accompanying supplementary material.}
    \label{fig:result}
\end{figure*}

\subsection{Same direction Between Ball's and Goalkeeper's trajectory}

Many goalkeepers in the top soccer league can actually predict the soccer ball's motion based on their years of experience. However, although the goalkeeper dived into the same direction as the soccer ball projected in, but they still did not prevent the ball from going into the goal frame. In that case, we should modified the motion of the goalkeeper slightly, so that the goalkeeper can stop the ball. 

Our approach can successfully correct the motion and help the goalkeeper stop the ball. We use the motion of the ball to optimize the motion of the goalkeeper. In figure \ref{fig:result}, it details how our animation is created. In the second example, we can find that the ball goes to the left hand side of the goalkeeper, and the goalkeeper jumps to left, too. However, the goalkeeper still misses the ball. In the third frame of the corrected animation, the goalkeeper catch the ball. Originally, in the forth row, the goalkeeper does not catch the ball, and the ball directly goes to the goal. 

We can also see that the motion is very similar, with the only real difference being how far the goalkeeper lunges, with the semantics of the jump kept very similar. This required 10 iterations of optimization.

\subsection{Opposite direction Between Ball's and Goalkeeper's trajectory}

These are the cases when the goalkeeper and ball are in opposite sides of motion which indicate that the goalkeeper had made the wrong judgement and dived in the wrong direction. This was the biggest challenge in our project because not only we had to completely correct the goalkeeper's direction but we also need to fix the goalkeeper's driving distance.

In order to fix the direction, we mirror the goalkeeper's joint position coordinates along the axis joining the head and the hip.
After this, we used our iterative optimization step which completely ensures the sufficient distance to dive through to save the ball. The problem is set up very similar to the subsection above and hence the method remains the same.

In the results seen in figure \ref{fig:result}, we can see the first 2 rows which have the motion mirrored (the second row and the third frame highlighted red), which ensures that the goalkeeper blocks the ball with the foot joint. Although this might be not the most ideal way to block the ball, it is still a somewhat feasible solution that obeying our problem statement and hypothesis which is the goalkeeper has to block the ball with any of the flexible joints and both the feet were defined so. While this highlights a limitation of our approach, the instance still accomplishes the task as desired. This required 10 iterations of the optimization.

\subsection{Goalkeeper has to not offer much/little movement}
This is the most peculiar and rarest of cases we had dealt with while rectifying the motion of the goalkeeper. For example, there is one case where the ball flies in a straight line into the middle of the goal frame. However, the goalkeeper gets confused and moves in the wrong direction due to the kicker's misleading hand gesture or footwork. The corrected motion for the goalkeeper to do is to stay still. In the third video that we analyzed, the goalkeeper moves in the right direction and has no particular dive motion and she should have rather stayed more firmly to her original position in frame 1.

This problem was tackled using the very same method which however requires the much lighter update on the motion coordinates of the goalkeeper. Since the virtual goal is nearly equivalent to the original video at that frame, the motion is nearly identical, but just translated a bit to ensure the goalkeeper intercepts the ball.

\section{Conclusion}
We present a novel technique of detecting and correcting the goalkeeper’s motion based on the ball’s motion while maintaining the semantic context of that motion. Our approach does not involve extensive use of machine learning while leveraging a modified version of a popular computer vision algorithm for keypoint detection, and interpolate coordinates independent of the frame size of the video. This method can be done in situations without complicated training algorithms since this method is based on mathematical concepts. We used an iterative algorithm for optimization of the manual/computer vision-based algorithm-based position detection for a ball as well as the goalkeeper. We contribute by presenting a well-balanced and nuanced algorithm that does not lean on exclusive machine learning usage and hence leads to a more real-time and adaptive approach.

\section{Limitation / Future work}
Although our approach can handle most of the cases which require correction and is pretty versatile, there are still some limitations. During the data collection phase, we had purposely chosen the videos that contained the front view of the penalty shot. Then, we can cut the video to have a front view only and be able to detect the goal. One of the possible projected solutions is that we can do prepossessing on the videos to transform the view to the front view.

For human detection, when the resolution is too low, we cannot detect the goalkeeper correctly which also applied to the ball detection. This limited our approach to a more manual supervision-based detection, unlike a regular computer vision-based. The possible solution to enhance our current detection technique is to create a custom automated ball detection, which adjusts the speed and the clarity of the ball using CV techniques. There are certain cases where our project is not very capable to detect humans in case there are many humans (even other players besides the goalkeeper and the scorer) in the video, there is a possible chance that the key point detection technique might select the wrong person as the goalkeeper. In these cases, we allow labeling the key points manually. We also encountered some significant challenges in determining the goalpost in the frames and the main reason we attribute was the lack of obtaining the 3D-based animation coordinates for all our motion coordinates.

Finally, as discussed in Section \ref{sec:results}, the model does not enforce feasible motion, and can sometimes make strange motion. In this case, the goalie slides to block the ball with their foot, which is somewhat uncommon. Other examples include the goalkeeper sliding their feet to meet the ball, rather than walking. Despite these issues, we are still able to produce robust results for this task, however this casts doubt onto the validity of applying this to more complex motions, and how to handle this remains an open research question. \\\\\\
\newpage
\newpage
\newpage
\newpage

\bibliographystyle{plain}
\bibliography{bibliography.bib}
\end{document}